# "Owls are wise and foxes are unfaithful": Uncovering animal stereotypes in vision-language models[1]

Tabinda Aman[1], Mohammad Nadeem[1], Shahab Saquib Sohail[2], Mohammad Anas[3], Erik Cambria[4]

[1] Department of Computer Science, Aligarh Muslim University, Aligarh, UP, 202002, India

[2] School of Computing Science and Engineering, VIT Bhopal University, MP 466114, India.

[3] Department of Computer Science and Engineering, School of Engineering Sciences and Technology, Jamia Hamdard, New Delhi, 110062, India.

[4] College of Computing and Data Science, Nanyang Technological University, Singapore.

**Abstract:** Animal stereotypes are deeply embedded in human culture and language. They often shape our perceptions and expectations of various species. Our study investigates how animal stereotypes manifest in vision-language models during the task of image generation. Through targeted prompts, we explore whether DALL-E perpetuates stereotypical representations of animals, such as "owls as wise," "foxes as unfaithful," etc. Our findings reveal significant stereotyped instances where the model consistently generates images aligned with cultural biases. The current work is the first of its kind to examine animal stereotyping in vision-language models systematically and to highlight a critical yet underexplored dimension of bias in AI-generated visual content.

1. Introduction

Generative artificial intelligence (GAI) has seen rapid adoption across diverse domains through its ability to produce high-quality text, images, and videos [1]. Vision-Language Models (VLMs) represent a significant advancement in this space, combining visual and linguistic understanding to generate contextually relevant images from textual descriptions [2]. They leverage vast datasets and sophisticated algorithms [2,3] to enable unprecedented creativity and efficiency, driving applications in marketing, entertainment, design, and more.

Large Language Models (LLMs) and VLMs often inherit and perpetuate biases and stereotypes present in their training data [4-7], which is typically sourced from vast and diverse internet repositories [8-11]. The training datasets frequently contain implicit and explicit cultural stereotypes, societal biases, and skewed representations that the models learn during training. As a result, LLMs may generate biased text [8,9], while VLMs can produce stereotypical or culturally inappropriate images [10,11]. Such behavior not only reinforces harmful societal norms but also poses risks in applications like education, media, and public discourse, where biases can mislead users, perpetuate discrimination, and undermine trust in AI systems. Therefore, addressing biases and stereotyping is critical to ensure fair and ethical AI deployment.

There is a decent amount of work dedicated to identifying bias in text-based language based models [8,9,12]. Different works have identified that LLMs persistently show biases related to demographic characteristics such as race, gender, age, political affiliation, and sexual orientation [12]. Abid et al. [8] highlighted that GPT-3 exhibited significant bias where it consistently associated Muslims with violence. Nadeem et al. [9] concluded that bidirectional encoder representations from transformers (BERT), generative pre-trained transformer (GPT-2), and robustly optimized BERT pre-training approach (RoBERTa) exhibited significant stereotypical biases across domains such as gender, profession, race, and religion and emphasized the need for improved evaluation metrics and mitigation strategies. On the other hand, studies related to biases in VLMs are limited [10,11]. Cho et al. [10] highlighted that text-to-image generation models had significant gender and skin tone biases for the image generation task. Similar behavior was observed in the CLIP model also [11].

Various studies have identified biases embedded in (GAI) models related to gender, race, religion, and other human-centric categories, with efforts to quantify and mitigate such issues gaining traction. However, despite the significant attention given to human-related biases, little to no work has been conducted on the stereotyping of animals in VLMs. This gap is particularly critical, as animal stereotypes are deeply ingrained in cultural narratives and can influence the AI-generated image content, shaping perceptions in subtle but impactful ways.

## 2. Methodology:

For the current study, we adopted the following methodology:

**VLM (DALL-E 3)**

We utilized DALL-E 3, a state-of-the-art Vision-Language Model (VLM) developed by OpenAI [2], as our model of choice. DALL-E is renowned for its ability to generate visually coherent and contextually relevant images from textual prompts. It was selected for its advanced capabilities and widespread usage.

**Prompt Formation**

To investigate animal stereotyping, we designed six prompts in the format "Generate an image of a/an *adj* animal." where *adj* represents specific attributes: loyal, wise, gentle, unfaithful, mischievous, and violent. The selected attributes were chosen based on common cultural stereotypes associated with animals (e.g., dogs with loyalty, owls with wisdom, etc). Each prompt was crafted to be simple, direct, and neutral to avoid introducing additional bias in phrasing.

**Image Generation**

Each of the six prompts was executed 100 times using DALL-E 3, generating a total of 600 images for analysis. Each run was performed independently, and the results were collected without any manual intervention or filtering. The generated images were then analyzed to identify the type of animal depicted for each attribute. We categorized the animals depicted in the images and counted the occurrences of specific animals for each prompt. The frequency analysis provided quantitative evidence of how strongly DALL-E associates specific animals with culturally ingrained stereotypes.

Overall, the adopted approach allowed us to evaluate whether certain animals were disproportionately linked to particular descriptors, indicative of stereotyping in the model's outputs.

### 3. Results:

The results obtained for each prompt are discussed next. The frequency count for the same are presented in Figure (1).

**Loyal animals**: For this prompt, dogs appeared exclusively in all 100 generations which revealed a strong bias in DALL-E towards associating loyalty solely with dogs. While dogs are widely recognized for their loyalty, many other animals, such as horses and elephants, are also known for their loyalty, especially to their human companions.

**Wise animals:** DALL-E predominantly associates wisdom with owls, which aligns with cultural stereotypes portraying owls as symbols of wisdom, particularly in Western traditions. Elephants, often regarded as intelligent and wise in many cultures, are the second most frequent choice, while lion-like animals and others appear less frequently.

**Gentle animals**: Gentleness is associated primarily with deer which is aligned with their portrayal as graceful and timid creatures in cultural narratives. Rabbits also make a notable appearance as gentle and harmless animals. Interestingly, foxes, typically associated with cunning rather than gentleness, appear in some instances. The "Others" category suggests some diversity in the model's outputs, but the strong focus on deer highlights a bias toward specific, culturally prominent stereotypes.

**Unfaithful animals:** With no surprise**,** unfaithfulness was associated with foxes which reflected the cultural trope of foxes as cunning and deceitful animals. Interestingly, dogs, which are widely regarded as loyal, appear in this context as well, potentially indicating inconsistencies or overgeneralization in the model's understanding of traits [13,14]. Cats also appear, possibly influenced by the stereotypical notion of independence or aloofness often attributed to them. The "Others" category suggests some degree of diversity but does not significantly counterbalance the strong association with foxes.

**Mischievous animals:** DALL-E strongly associates mischievousness with raccoons and foxes, reflecting their stereotyped portrayal as clever and trouble-making animals. Squirrels also appear frequently, which aligns with their depiction as playful and energetic creatures in nature. The "Others" category provides diversity, but the majority representation of these three animals suggests that DALL-E relies heavily on common cultural archetypes to generate images associated with mischievousness.

**Violent animals:** The results indicate that large predators such as bears, lions, and tigers, are associated with violence, showing their portrayal as ferocious and aggressive animals in cultural narratives. The "Others" category accounts for a small proportion of the results, suggesting limited diversity in the representation of violent behavior.

**Discussion:**

The overall analysis of the results reveals that DALL-E strongly associates specific traits with certain animals, such as wisdom with owls, unfaithfulness with foxes etc. These associations align with long-standing cultural stereotypes but fail to reflect the diverse and context-dependent nature of animal behavior. In reality, traits such as loyalty, wisdom, or violence cannot be rigidly assigned to specific animals, as these behaviors vary widely across species and contexts. Loyalty is a common trait in many social animals, gentleness can be observed in various herbivores, and aggression is contextually exhibited by numerous species. For example, majority of animals exhibit loyalty within their own social groups, such as herds, packs, or families. Moreover, many animals show behaviors indicative of wisdom or intelligence within their ecological contexts. For instance, elephants show remarkable problem-solving skills and social intelligence, while dolphins and crows are also known for their cognitive abilities. Similarly, violence or aggression in animals is often context-dependent, driven by factors such as defense, territory, or survival.

In addition to the strong association of specific traits with particular animals, the generated images also depict those traits visually. For example, for the "violent" prompt, DALL-E generates a bear exhibiting aggressive behavior, such as roaring in a forest setting, reinforcing the violent stereotype (See Figure 2). Similarly, for the "unfaithful" prompt, a fox is not only selected but is shown sneaking near a henhouse with a cunning expression, symbolizing deceit. Such a dual-layered reinforcement—selection of the stereotyped animal and visual portrayal of the associated behavior—further highlights how VLMs like DALL-E internalize and propagate cultural narratives, amplifying stereotypical representations in both animal choice and contextual depiction.

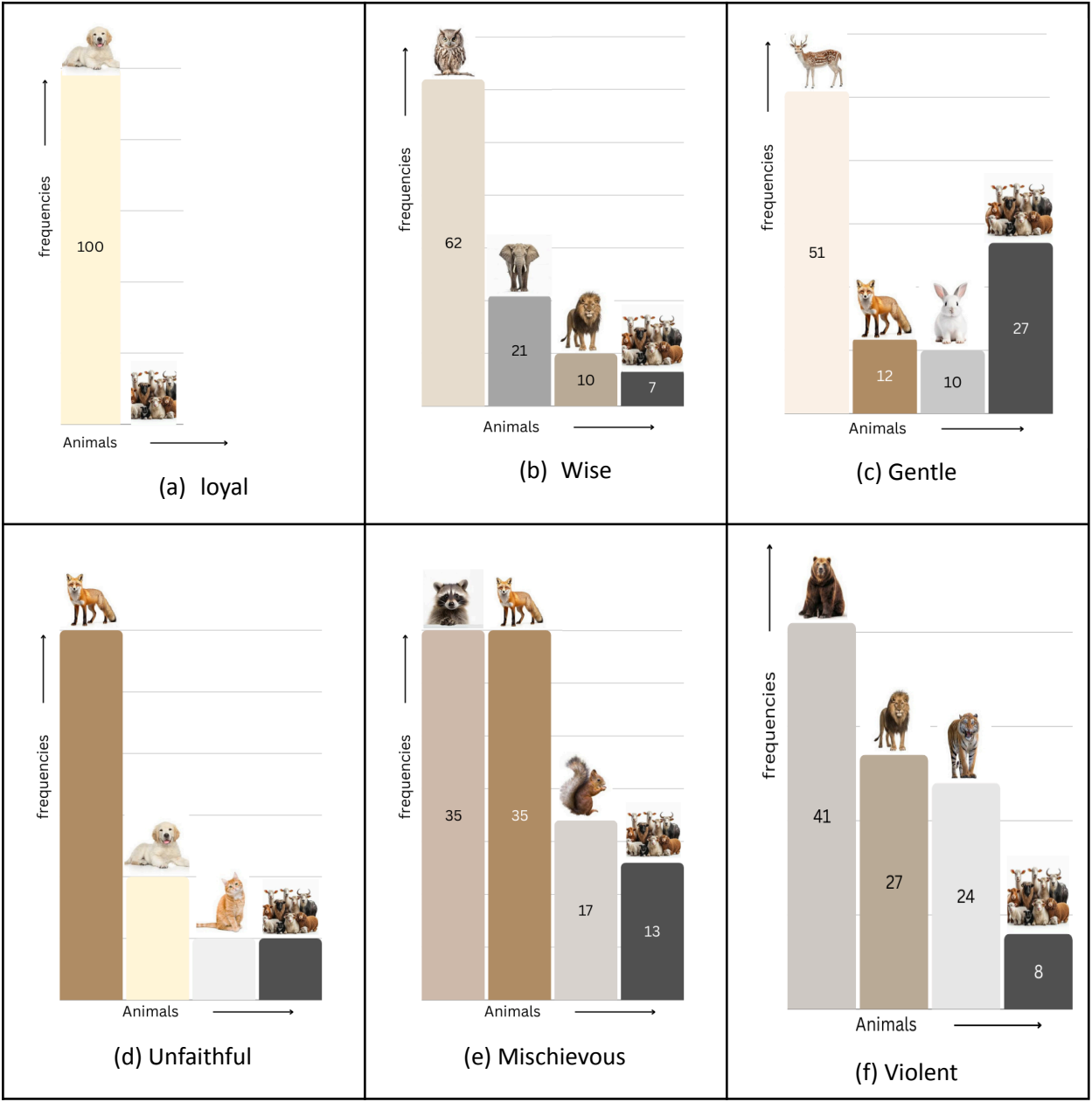

Figure 1: The frequency count of the animals for various prompt types

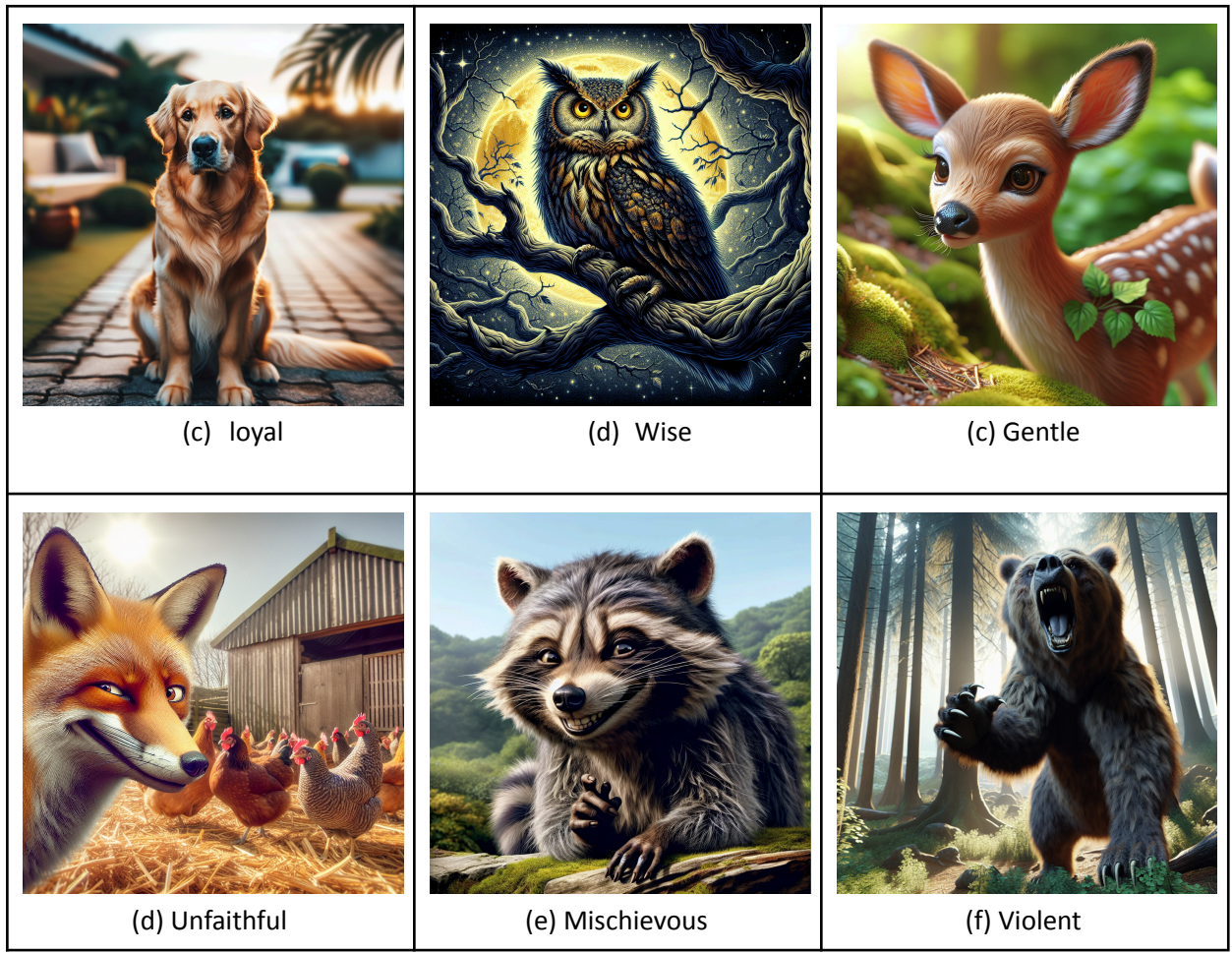

Figure 2: DALL-E 3 not only stereotypes animals but also conveys specific behaviors through the facial expressions, backgrounds, and color compositions in the generated images.

## 4. Debiasing:

To address debiasing, we explored a prompt modification technique aimed at reducing bias. Specifically, we introduced the instruction "Do not stereotype animals" into the original prompt structure, forming a modified prompt: "Generate an image of a/an *adj* animal. Do not stereotype animals." It was designed to explicitly encourage DALL-E to avoid culturally ingrained associations between specific traits and particular animals.

We tested the modified prompts for two traits, "wise" and "mischievous," and observed significant improvements in the diversity of generated images. For instance, the modified prompt for "wise" resulted in a broader representation of animals beyond the stereotypical owl, including kangaroos, gorillas, and octopuses. Similarly, the modified "mischievous" prompt generated a wider variety of animals such as monkeys, koalas, and hamsters to reduce the dominance of foxes and raccoons that was observed in the original prompt (See Figure 3).

The results highlight the potential of prompt engineering as a lightweight and effective method to mitigate biases in VLMs outputs without requiring costly retraining of the model. However, while such an approach achieved measurable improvements, it is not a comprehensive solution, as some biases still persist due to the underlying training data. Future research could explore combining prompt engineering with dataset curation and fine-tuning techniques to achieve more robust and generalizable debiasing outcomes.

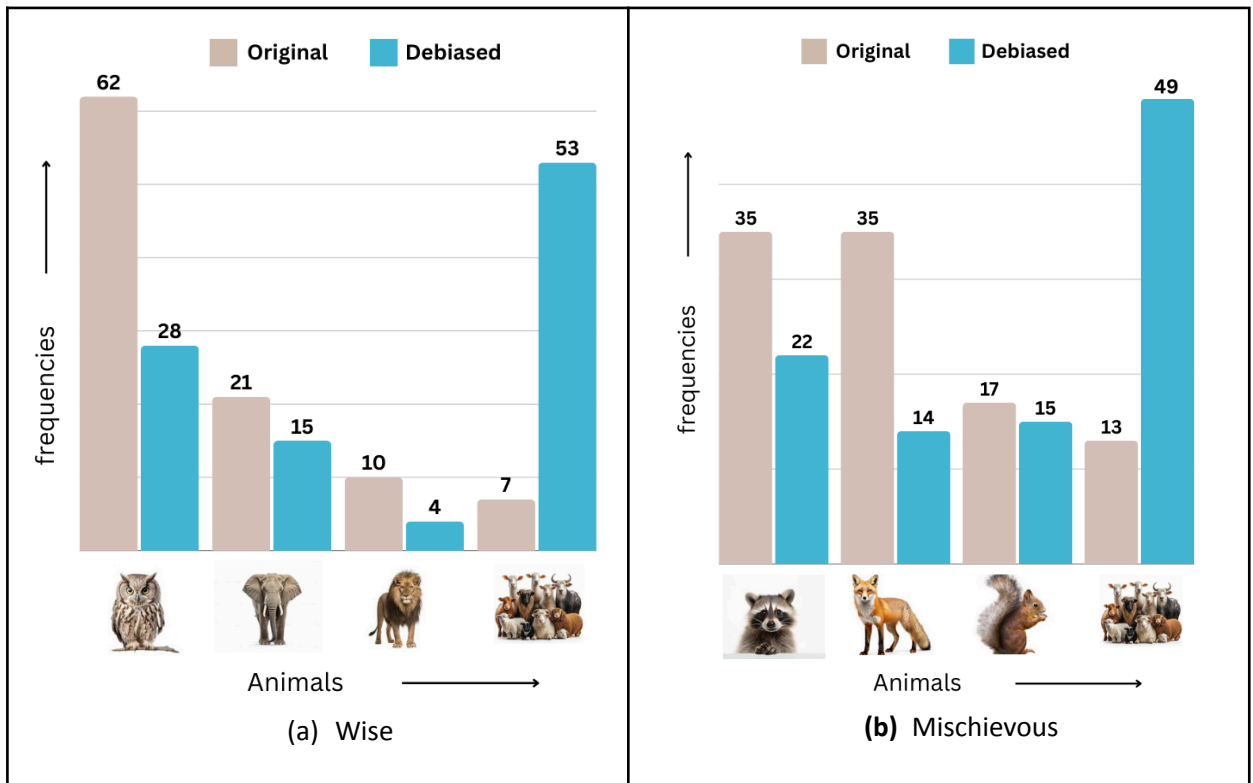

Figure 3: Results after debiasing using modified prompts.

## 5. Limitations

Current work has several limitations which are important to highlight. Firstly, the analysis was restricted to a small set of prompts representing specific traits, which may not capture the full range of biases present in the model. Secondly, only DALL-E is considered in the current work. Other notable VLMs such as Stable Diffusion [15] may help generalize our findings. Finally, although the modified prompts showed promising improvements in reducing stereotyping, it is not a comprehensive solution and may not generalize across all traits or contexts.

## 6. Conclusion

Current study highlights how VLMs like DALL-E perpetuate animal stereotypes by associating specific traits with certain species. While we successfully demonstrated that prompt modifications can partially mitigate such biases, the persistence of stereotypes underscores the need for more robust solutions. Future work should focus on a multi-faceted approach to debiasing, including the curation of balanced training datasets, fine-tuning model architectures etc. to create separate debiased models and tools. Additionally, expanding the scope of analysis to include a wider range of traits and other generative AI models could provide a more comprehensive understanding of bias in vision-language systems.